\documentclass{article}
\usepackage[utf8]{inputenc}
\usepackage{amsmath,amssymb,amsthm}
\usepackage{tikz}
\usetikzlibrary{decorations.markings,arrows,automata,arrows,backgrounds,snakes}
\usepackage{graphicx}
\usepackage{caption}
\usepackage{subcaption}

\usepackage{tikz}
\usetikzlibrary{shapes,arrows}
\usepackage{verbatim}

\newtheorem{thm}{Theorem}[subsection]

\newtheorem{lem}[thm]{Lemma}
\newtheorem{rem}[thm]{Remark}

\theoremstyle{definition}
\newtheorem{defn}[thm]{Definition}
\usepackage[english]{babel}
\usepackage[nottoc]{tocbibind}
\allowdisplaybreaks

\title{Theory of periodic convolutional neural network}
\author{Yuqing Liu \\
Department of Mathematics, HKU
 \\
Email: yuliu6@hku.hk\\
lyqing199669@gmail.com
}
\date{Oct 2025}

\begin{document}

\maketitle
\begin{abstract}
We introduce a novel convolutional neural network architecture, termed the \emph{periodic CNN}, which incorporates periodic boundary conditions into the convolutional layers. Our main theoretical contribution is a rigorous approximation theorem: periodic CNNs can approximate ridge functions depending on $d-1$ linear variables in a $d$-dimensional input space, while such approximation is impossible in lower-dimensional ridge settings ($d-2$ or fewer variables). This result establishes a sharp characterization of the expressive power of periodic CNNs. Beyond the theory, our findings suggest that periodic CNNs are particularly well-suited for problems where data naturally admits a ridge-like structure of high intrinsic dimension, such as image analysis on wrapped domains, physics-informed learning, and materials science. The work thus both expands the mathematical foundation of CNN approximation theory and highlights a class of architectures with surprising and practically relevant approximation capabilities.
\end{abstract}

\section{Introduction}

Convolutional neural networks (CNNs) form the backbone of modern machine learning for high-dimensional structured data, especially images, videos, and scientific signals. Despite their empirical success, the theoretical understanding of their approximation properties remains incomplete. A central question is: \emph{what classes of functions can specific CNN architectures approximate efficiently?}

In this work, we focus on a new architectural variant we call the \emph{periodic CNN}. The key modification is the enforcement of periodic boundary conditions in convolutional operations. Such a structure arises naturally in many scientific and engineering applications, for example, when signals live on tori, in materials with repeating microstructure, or in spectral data analysis.

Our main result is a sharp approximation theorem: \textbf{periodic CNNs can approximate ridge functions depending on $d-1$ linear directions in $d$ dimensions, but not ridge functions depending on $d-2$ or fewer directions.} Ridge functions, defined as $f(x) = g(a^\top x)$ with a univariate profile $g$, play a central role in approximation theory and are widely used to characterize the expressive limitations of neural networks. The result demonstrates that periodic CNNs occupy a distinctive position in the approximation hierarchy: they are powerful enough to capture nearly full-dimensional ridge structures, yet strictly limited against lower-dimensional ones.

This dichotomy yields both \emph{theoretical insights and practical implications}. On the theory side, it provides one of the first precise separation results for CNN variants, showing exactly where periodicity enables expressivity and where it constrains it. On the practical side, it suggests that periodic CNNs are particularly well matched to high-dimensional structured signals where information is concentrated along $d-1$ directions, a common scenario in physics-informed learning and scientific computing.

The remainder of the paper is organized as follows. Section 2 defines the periodic CNN architecture formally. Section 3 establishes the approximation theorem for $d-1$-dimensional ridge functions and proves the impossibility result for $d-2$ dimensions and below. Section 4 discusses implications for applications and outlines future directions.

Ridge functions $f(x) = \phi(a^\top x)$ are a cornerstone of approximation theory, since they capture the simplest form of high-dimensional functions depending only on low-dimensional projections. 
They have been widely studied in numerical analysis, harmonic analysis, and machine learning (see, e.g., \cite{pinkus2015ridge, barron1993universal}), both as hard test cases for approximation lower bounds and as building blocks for constructive approximation schemes.
In practice, ridge-shaped structures appear in regression with latent variables, signal remixing, and physics-informed learning where dominant directions of variation exist.

\section{Periodic CNN Architecture}
Before analyzing approximation properties, we carefully describe the class of networks under study. 
The periodic CNN is obtained by replacing the usual finite-support convolutional operators with their circular analogs, i.e., convolutions performed under periodic boundary conditions. 
This modification equips the network with a natural translation-equivariant structure on the torus $\mathbb{T}^d$, aligning the architecture with many scientific settings where data are inherently periodic (e.g., spectral signals, crystallographic structures, or functions defined on wrapped domains). 

In this section, we formalize the definitions of convolution, activation, and network hypothesis space under the periodic setting, laying the groundwork for the approximation theorems in Section~3.
\subsection{Prelims}
\noindent
In this section, we introduce the deep CNN structure with its mathematical definition.\\
\noindent
In our content, we consider the deep CNNs with two essential parts: the ReLU activation function(which is a univariate nonlinear function) given by

\begin{center}
\begin{equation}
     \sigma(x)=(x)_{+}=\max\{x,0\}, x\in \mathbb{R}
\end{equation}
\end{center}
and an infinite sequence of filter $\mathbf{w}= \{ w^{(j)}\}_{j}$ which would induce sparse convolutional structures. The filter $w=(w_{k})_{k=-\infty}^{\infty}$ is a sequence of coefficient. In addition, we assume $w_{k}^{(j)} \neq 0, 0 \leq k \leq s$ and $s \geq 2$. The convolution of $w$ with another sequence $p=(p_1, \cdots ,p_{D})$ denote as $w*p$, which is defined as
\noindent
\begin{center}
    \begin{equation}
    (w\ast p)_{i}= \sum_{k=0}^{D} w_{i-k}p_{k}
\end{equation}
\end{center}
which gives a $D\times D$ Toeplitz convolutional matrix T with constant diagonals 
\medskip

\begin{equation}
    T = \begin{bmatrix}
    w_{0} & 0 & 0 &\cdots & w_{s} & \cdots &w_{2} &w_{1}\\
    w_{1} & w_{0}& 0&\cdots & \cdots  & w_{s}&\cdots &w_{2}\\
    \vdots & \cdots &\ddots&\ddots &\ddots & \ddots& \ddots&\vdots\\
    w_{s-1} & w_{s-2}& \cdots&w_{0} &\cdots & \ddots& \ddots &w_{s}\\
    w_{s} & w_{s-1}& \cdots&0 &\ddots & \ddots& \ddots &0\\
    0 & w_{s}& \cdots&0 & \cdots & \ddots& \ddots &0\\  
    \vdots &\ddots& \ddots&\vdots & \cdots & \cdots& \ddots &\vdots\\
     \vdots &\cdots& \cdots&\ddots & \cdots & \cdots& \cdots &\vdots\\
     0 &\cdots& \ddots&\vdots &  w_{s} & \cdots& w_{0} &w_{1}\\
    0 & \cdots& \cdots&\cdots &\cdots& w_{s}& \cdots &w_0{}\\
    
    \end{bmatrix}
\end{equation}
\medskip
\noindent

We denote $\mathbb{T} := \mathbb{R}/\mathbb{Z}$ for the unit circle, identified with the interval $[0,1)$ with periodic boundary conditions. 
The $d$-dimensional torus is $\mathbb{T}^d = (\mathbb{R}/\mathbb{Z})^d$, equipped with the normalized Lebesgue measure. 
Functions on $\mathbb{T}^d$ thus admit Fourier expansions
\[
f(x) = \sum_{k\in\mathbb{Z}^d} \widehat f(k)\,e^{2\pi i k\cdot x},
\]
with Parseval identity $\|f\|_{L^2(\mathbb{T}^d)}^2 = \sum_{k\in\mathbb{Z}^d}|\widehat f(k)|^2$.

\subsection{Periodic CNN}
\begin{defn}\label{sec:definition}
Given an input vector $x \in \mathbb{R}^{s}$, a deep convolutional neural network (CNN) of depth $J$ is defined as a sequence of feature vectors $\{h^{(j)}(x)\}_{j=0}^J$ with initialization
\[
h^{(0)}(x) = x,
\]
and recursive update
\begin{equation}
    h^{(j)}(x) = \sigma\!\big( T^{(j)} h^{(j-1)}(x) - b^{(j)} \big), 
    \quad j = 1,2,\ldots,J,
\end{equation}
where
\begin{itemize}
    \item $T^{(j)} = \big( w_{i-k}^{(j)} \big)$ denotes a $d \times d$ \emph{circular convolutional matrix}, i.e., convolution with periodic boundary conditions;
    \item $\sigma$ denotes the ReLU activation function acting componentwise;
    \item $b^{(j)} \in \mathbb{R}^{d_j}$ denotes the bias vector at layer $j$.
\end{itemize}

To emphasize periodicity, we restrict the bias vector $b^{(j)}$ to have the special form
\begin{equation}
    b^{(j)} = 
    \big[ b_{1}, \ldots, b_{s}, 
    \underbrace{b_{s+1}, \ldots, b_{s+1}}_{d_j - 2s \ \text{times}}, 
    b_{d_j - s+1}, \ldots, b_{d_j} \big]^{\!\top},
\end{equation}
i.e., with $d_j - 2s$ repeated components in the middle.
We refer to the resulting network as a \emph{periodic CNN}. 
\end{defn}

Thus, periodic CNNs are precisely standard CNNs except that all convolutions are defined with periodic boundary conditions, making the network architecture equivariant with respect to translations on the torus $\mathbb{T}^d$.

In order to characterize the class of functions representable by periodic CNNs, we next introduce the associated hypothesis space.

\begin{defn}\label{space}
For the periodic CNN defined in Definition~\ref{sec:definition}, the hypothesis space is the set of functions of the form
\begin{equation}
    \mathcal{H}^{\mathbf{w},\mathbf{b}}
    = \Big\{ f(x) = \sum_{k=1}^{d_{J}} c_{k} \, h_{k}^{(J)}(x) 
    : c \in \mathbb{R}^{d_{J}} \Big\},
\end{equation}
where $h^{(J)}(x)$ denotes the final feature vector produced by the depth-$J$ periodic CNN, and 
$\mathbf{w} = \{w^{(j)}\}_{j=1}^J$, $\mathbf{b} = \{b^{(j)}\}_{j=1}^J$ denote the collection of weights and bias vectors, respectively.
\end{defn}

\section{Main Result}
We now turn to the central theoretical question: 
What functions can periodic CNNs approximate, and where do their limitations lie? 
Our analysis reveals a sharp dichotomy in terms of ridge functions. 
On the one hand, periodic CNNs can successfully approximate ridge functions depending on $(d-1)$ linear directions, capturing nearly full-dimensional ridge structure. 
On the other hand, they fundamentally fail to approximate ridge functions depending on $(d-2)$ or fewer directions, no matter how large the depth or width of the network.

This section is organized into three parts. 
First, we prove the constructive approximation theorem for $(d-1)$-dimensional ridge functions. 
Second, we establish the impossibility result for $(d-2)$-dimensional ridge functions. 
Finally, we provide a concrete counterexample that illustrates the impossibility argument in an explicit Fourier calculation.
\subsection{Approximation of $(d-1)$-dimensional ridge functions}
We first establish the positive approximation result. 
In this subsection, we prove that periodic CNNs are sufficiently expressive to approximate ridge functions depending on $(d-1)$ linear directions. 
This constructive theorem highlights the near-full-dimensional power of the architecture.

\begin{thm}
Given weight vectors $w_{1}, w_{2}, \cdots w_{J}$ and bias vectors $b_{1}, b_{2}, \cdots b_{J}$, and let $W \cdot x = y$, where $W \in \mathbb{R}^{d}$, $x \in \mathbb{R}^{d}$. \\
We define the relation $\preceq$ such that 
\begin{equation}
    (w_1,b_1) \preceq (w_2,b_2) 
\end{equation}
if and only if
\begin{equation}
    w_{1} \cdot y - b_1 > 0 \Rightarrow   w_{2} \cdot y - b_2 > 0, \qquad \forall y \in \mathbb{R}
\end{equation}
$\Longleftrightarrow$\\

\begin{equation}
    y > \frac{b_1}{w_1} \Rightarrow  y > \frac{b_2}{w_2}
\end{equation}
\end{thm}

\noindent
\medskip
\begin{lem}cf.~\cite{zhou2019universality}
Let $s \geq 2$ and $W = (W_{k})_{- \infty}^{\infty}$
be a sequence supported in ${0, ...,M}$ with $M \geq 0$. Then there
exists a finite sequence of filter masks $\{w^{(j)}\}^{J}_{j=1}$
supported in ${0, ..., s}$ with $J < \frac{ M}{s-1 }+ 1$ such that the
convolutional factorization $W = w^{(J)}\ast ... \ast w^{(2)} \ast w^{(1)}$ holds true.
\end{lem}
\noindent
\medskip

\begin{thm}
Let $2 \leq s \leq d$. Given a ridge function $f(a\cdot x): \mathbb{R}^{d} \longrightarrow \mathbb{R}$, where $f$ is a H\"older $\alpha$ function for some $\alpha > 0 $.  Then there exists a deep network stated in Definition 1.2.1 with $J$ layer, $\{w^{i}\}_{i=1}^{J}$, $\{b^{i}\}_{i=1}^{J}$ such that

\begin{equation}
    \lim_{J \to \infty}\{ ||c \cdot h^{(J)} - f(a \cdot \circ)|| \}_{C(\Omega)} = 0
\end{equation}
\end{thm}

\hfill
\begin{proof}
We proceed with the proof in the following steps\\

Step1: We consider an infinite circular vector $$\textbf{x}=\begin{bmatrix}
     \vdots\\
       x_{d}\\
       x_{1}\\
       \vdots\\
       x_{d}\\
       x_{1}\\
       \vdots\\     
\end{bmatrix} $$
by equation (2), the convolution is given by\\
 $$w^{(1) }\ast \textbf{x} = w^{(1) }\ast \begin{bmatrix}
       \vdots\\
       x_{d}\\
       x_{1}\\
       \vdots\\
       x_{d}\\
       x_{1}\\
       \vdots\\
\end{bmatrix} = \begin{bmatrix}
         \vdots \\
         
       w_{d-1}^{(1)}x_{1} +\cdots + w_{D-d}^{(1)}x_{d}\\
    
       w_{0}^{(1)}x_{1}+\cdots +w_{D-d+1}^{(1)}x_{d}  \\
      
       \vdots\\
       w_{d-1}^{(1)}x_{1} +\cdots + w_{D-d}^{(1)}x_{d}\\
       
       w_{0}^{(1)}x_{1}+\cdots +w_{D-d+1}^{(1)}x_{d} \\
       \vdots\\
\end{bmatrix}$$\\

The output of the first convolution layer is the first $d$ terms as well as the bias terms

 $$h^{(1)}(x)= \sigma ([w^{(1) }\ast \textbf{x}]_1^d -b^{(1)}) = \sigma\left(\begin{bmatrix}
         
       w_{0}^{(1)}x_{1}+\cdots +w_{D-d+1}^{(1)}x_{d} -b^{(1)}_1\\
      
       w_{1}^{(1)}x_{1}+\cdots +w_{D-d+2}^{(1)}x_{d} -b^{(1)}_2\\
       \vdots\\
      w_{d-1}^{(1)}x_{1} +\cdots + w_{D-d}^{(1)}x_{d}-b^{(1)}_d\\
     
\end{bmatrix}\right),$$\\
where $[z]_{1}^{n}=[z_{1},\cdots z_{n}]^{T}$ for a sequence $z=(z_{k})_{k=-\infty}^{\infty}$\\

Now, we consider \\
\begin{equation*}
    w^{(2)} \ast \begin{bmatrix}
    \vdots\\
    (w^{(1)} \ast x)_{d}\\
    (w^{(1)} \ast x)_{1}\\    
    \vdots\\
    (w^{(1)} \ast x)_{d}\\
    \vdots
\end{bmatrix} =  w^{(2)} \ast (w^{(1) }\ast \textbf{x})= \sum_{i+t=n}w^{(2)}_{i}(w^{(1)}_{j} \ast x_{k})_{t}
\end{equation*}
Where the summation can be represented as\\

\begin{equation*}
\begin{split}
   &\sum_{i+t=n}w^{(2)}_{i}(w^{(1)}_{j} \ast x_{k})_{t}= \sum_{\substack{i+t=n \\ t\leq 0}}w^{(2)}_{i}(w^{(1)}_{j} \ast x_{k})_{t}+ \sum_{\substack{i+t=n \\ t > 0}}w^{(2)}_{i}(w^{(1)}_{j} \ast x_{k})_{t}\\
   & =\sum_{\substack{i+j+k=n \\j+k \leq 0}}w^{(2)}_{i}w^{(1)}_{j}x_{k} +\sum_{\substack{i+j+k=n \\j+k > 0}}w^{(2)}_{i}w^{(1)}_{j}x_{k} = \sum_{i+j+k=n }w^{(2)}_{i}w^{(1)}_{j}x_{k}
   \end{split}
\end{equation*}
\\

On the other hand,
\begin{equation*}
    (w^{(2)} \ast w^{(1)}) \ast \begin{bmatrix}
             \vdots\\
       x_{d}\\
       x_{1}\\
       \vdots\\
       x_{d}\\
       \vdots\\     
    \end{bmatrix} =  (w^{(2)} \ast w^{(1) })\ast \textbf{x} = \sum_{i+t=n} (w^{(2)}w^{(1)})_{t} \ast x_{k} =  \sum_{i+t=n} w^{(2)}_{i}w^{(1)}_{j}x_{k}
\end{equation*} 
\\
Therefore,\\
\begin{equation*}
    w^{(2)} \ast \begin{bmatrix}
    \vdots\\
    (w^{(1)} \ast x)_{d}\\
    (w^{(1)} \ast x)_{1}\\    
    \vdots\\
    (w^{(1)} \ast x)_{d}\\
    \vdots
\end{bmatrix}  =  (w^{(2)} \ast w^{(1)}) \ast \begin{bmatrix}
             \vdots\\
       x_{d}\\
       x_{1}\\
       \vdots\\
       x_{d}\\
       \vdots\\     
    \end{bmatrix}
\end{equation*}
\\
Choose $b^{(1)}$ to be small enough such that $(w^{(1)}\ast \textbf{x}) - b^{(1)} \geq 0 $. The second convolutional layer is\\
\begin{equation*}
\begin{split}
    h^{(2)} &= \sigma(w^{(2)}\ast \sigma (([w^{(1) }\ast \textbf{x}]_1^d ) -b^{(1)})-b^{(2)})\\
    &=\sigma ( [w^{(2)} \ast w^{(1)} \ast \textbf{x} ]_1^d - B^{(2)} ,
\end{split}
\end{equation*}\\
\medskip
where $B^{(2)} =( w^{2} \ast b^{(1)})+b^{(2)} $\\
Inductively, we show that the $J_1-th$ convolutional layer is \\
\begin{equation*}
     h^{(J_1)} = \sigma ( [w^{(J)} \ast \cdots \ast w^{(1)} \ast \textbf{x} ]_1^d - B^{(J)})
\end{equation*}
where $B^{(J_1)} $ is the bias at the $J_1-th$ layer
\medskip

Step 2:
By lemma 2.0.2, we have $W=w^{(J_1)}\ast ... \ast w^{(2)} \ast w^{(1)}$, where $W$ is supported on $\{0, \cdots, d-1\}$, i.e, $W=[W_{d-1}, \cdots W_{0}, 0, \cdots]$\\
By equation $(2)$, the convolution $W \ast x$ can be represented as\\
\begin{equation}
\begin{bmatrix}
        W_{0} & 0 & 0 &\cdots & W_{d-1} & \cdots  &W_{1}\\
    W_{1} & W_{0}& 0&\cdots & \cdots &\cdots &W_{2}\\
    \vdots & \cdots &\ddots&\ddots &\ddots & \ddots&\vdots\\
    W_{s} & W_{s-1}& \cdots&W_{0} &\cdots & \ddots &W_{s+1}\\
    \vdots &\ddots& \ddots&\cdots & \ddots & \ddots &\vdots\\
    W_{d-1} & W_{d-2}& \cdots&0 & \cdots & \ddots &0\\  
    \vdots &\ddots& \ddots&\vdots & \cdots & \ddots &\vdots\\
    0 & \cdots& \cdots&\cdots &\cdots& W_{0}
    &W_{d-1}\\
    0 & \cdots& \cdots&\cdots &\cdots& \cdots
    &W_{0} \\
\end{bmatrix}\cdot x
\end{equation}\\

Observe that the $d-th$ row is\\
$(W \ast x)_{d-1}=[W_{d-1}, \cdots W_{0}, 0, \cdots] \cdot x = W_{d-1} \cdot x_1+\cdots W_{0}\cdot x_{d}=W\cdot x$\\
\medskip \\
Step 3:
Let $y = W \cdot x$, where $[W_{d-1}, \cdots, W_{0}] = [a_{1}, \cdots , a_{d}] $. \\
Then for any $\epsilon > 0$, there exists sequences $\textbf{w}$ of filter masks, $\textbf{b}$ of bias vectors and $\sum_{j=1}^{J_2}\sigma(w_{j} \cdot y -b_{j})$, such that \\
\begin{equation}
    |\sum_{j=1}^{J_1}\sigma(w_{j} \cdot y -b_{j}) - f(y)| < \epsilon \qquad \text{ for all } y
\end{equation}\\
Without loss of generality, assume 
\begin{equation}\label{preceq}
    (w_1, b_1) \preceq (w_2, b_2) \preceq \cdots \preceq (w_{J_2}, b_{J_2})
\end{equation}
Choose the bias vector $B^{(J_{1})}$ at layer $J_{1}$ such that the $J_{1}-th$ layer is given by\\
\begin{equation*}
    \begin{bmatrix}
            0\\
            \vdots\\
            0\\
            y+A\\
    \end{bmatrix}
\end{equation*}
with $A$ be a real number such that $a\cdot x+A\geq0$ for all $x$.\\\medskip
Let the weights vectors at the $(J_{1}+1)-th$ layer be\\
\begin{equation*}
w^{(J+1)}=\begin{bmatrix}
    w_{1}\\
    1\\
    0\\
    \vdots\\
    0
\end{bmatrix}
\end{equation*}
and choose the bias vectors at the $(J_1+1)-th$ layer to be\\
\begin{equation*}
B^{(J_1+1)}=\begin{bmatrix}
    0\\
    0\\
    0\\
    \vdots\\
    w_{1}A+b_1
\end{bmatrix}.
\end{equation*}
Then we have \\
\begin{equation*}
    h^{(J_1+1)}(x) = \sigma(h^{(J_1)}(x) \ast w^{(J_1+1)} -B^{(J_1+1)}) = \begin{bmatrix}
            y+A\\
            \vdots\\
            0\\
           \sigma (w_{1}y-b_1)
    \end{bmatrix} 
\end{equation*}\\
\medskip
Similarly, Let the weight vectors at the $(J_1+2)-th$ layer to be\\ 
\begin{equation*}
w^{(J_1+2)}=\begin{bmatrix}
    w_{2}\\
    1\\
    0\\
    \vdots\\
    0
\end{bmatrix}
\end{equation*}
and choose the bias vectors at the $(J_1+2)-th$ layer to be\\
\begin{equation*}
B^{(J_1+2)}=\begin{bmatrix}
     w_{2}A+b_2\\
    0\\
    0\\
    \vdots\\
   \sigma (w_{1}y-b_1)
\end{bmatrix}.
\end{equation*}
\medskip
Then we have \\
\begin{equation*}
    h^{(J_1+2)}(x) = \sigma( h^{(J_1+1)}(x)  \ast w^{(J_1+2)} -B^{(J_1+2)}) = \begin{bmatrix}
            \sigma(w_2y-b_2+\sigma(w_1y-b_1))\\
            y+A\\
            0\\
            \vdots\\
            0\\
    \end{bmatrix} 
\end{equation*}\\
\medskip
Inductively, we would have that the $(J_1+J_2)-th$ convolutional layer would be\\
\begin{equation}
   h^{(J_1+J_2)}(x) = \begin{bmatrix}
            0\\
            \vdots\\
            0\\
            \sigma (\cdots \sigma(\sigma(w_{1}\cdot y -b_{1}) + w_{2} \cdot y - b_{2})\cdots+w_{J_2}y- b_{J_2})\\
            y+A\\
            0\\
            \vdots\\
            0\\
    \end{bmatrix}\in\mathbb{R}^d,
\end{equation}\\\medskip
where $y+A=(h^{(J_1+J_2)}(x))_{J_2\mod{d}}$.\\

Consider
$$\sigma (\cdots \sigma(\sigma(w_{1}\cdot y -b_{1}) + w_{2} \cdot y - b_{2})\cdots+w_{J_2}y- b_{J_2}).$$\\
We show it can be written as a summation of form $\sigma(w_{j}y-b_{j})$.\\
Assume 
\begin{equation*}
    \sigma(\cdots \sigma(w_1y-b_1)\cdots+w_{k}y-b_{k}) = \sum_{j=1}^{k} \sigma(w_{j}y-b_{j})
\end{equation*}
Since (\ref{preceq}), we have
\begin{equation*}
\begin{split}
     & \sigma(\cdots \sigma(w_1y-b_1)\cdots+w_{k+1}y-b_{k+1}) \\
     & = \sigma(\sum_{j=1}^{k} \sigma(w_{j}y-b_{j})+w_{k+1}y-b_{k+1})\\
     & =  \left\{
\begin{array}{ll}
    \sigma(w_{k+1}y-b_{k+1}), \qquad \qquad  \qquad &\text{if } w_{j}y-b_{j} \leq 0, j=1,\cdots,k \\
    \sigma(\sum_{j=j'}^{k} \sigma(w_{j}y-b_{j}) + w_{k+1}y-b_{k+1} ), \qquad  &\text{if } w_{j'}y-b_{j'} > 0 \text{ for some }j'=j
\end{array} 
\right. 
\end{split}
\end{equation*}\\
\medskip
In both cases, we have\\
\begin{equation*}
     \sigma(\cdots \sigma(w_1y-b_1)\cdots+w_{k}y-b_{k}) = \sum_{j=1}^{k+1} \sigma(w_{j}y-b_{j}) 
\end{equation*}
Thus inductively, we have\\
\begin{equation}
  \sigma (\cdots \sigma(\sigma(w_{1}\cdot y -b_{1}) + w_{2} \cdot y - b_{2})\cdots+w_{J_2}y- b_{J_2})=  \sum_{j=1}^{J_2}\sigma(w_{j} \cdot y -b_{j})
\end{equation}
Thus, we can apply Theorem 1 in \cite{Zhou} and thus complete the proof

\end{proof}

\hfill

\subsection{Impossibility for $(d-2)$-dimensional ridge functions}
Having established the positive case, we now turn to the corresponding limitation. 
We show that if the ridge profile depends on at most $(d-2)$ coordinates, then the periodic CNN hypothesis space cannot approximate the target function in $L^2$ to arbitrary accuracy. 
This impossibility result forms the sharp counterpart to the previous theorem.

\begin{defn}[Network lattice and associated subspace]\label{def:Lambda_H}
Let the convolutional filters (stencils) used by the periodic CNN have finite support sets
\(K_1,\dots,K_J\subset\mathbb{Z}^d\). Define the additive subgroup (lattice)
\[
\Lambda \;:=\; \big\langle K_1\cup\cdots\cup K_J\big\rangle_{\mathbb{Z}}
\;=\;\Big\{ \sum_{j} n_j k_j : n_j\in\mathbb{Z},\ k_j\in \bigcup_{i=1}^J K_i \Big\}\subset\mathbb{Z}^d.
\]
Denote by \(H_\Lambda\) the closed subspace of \(L^2(\mathbb{T}^d)\) spanned by exponentials with frequencies in \(\Lambda\):
\[
H_\Lambda := \overline{\operatorname{span}}\{ e^{2\pi i k\cdot x} : k\in\Lambda\}^{\,L^2(\mathbb{T}^d)}.
\]
\end{defn}

\begin{lem}[Fourier representation of a ridge function]\label{ridge}
Let $\phi:\mathbb{T}\to\mathbb{C}$ have Fourier series
\[
\phi(t) \;=\; \sum_{m\in\mathbb{Z}} c_m \, e^{2\pi i m t}, \qquad c_m\in\mathbb{C}.
\]
Fix $u\in\mathbb{R}^d$ and define $f:\mathbb{T}^d\to\mathbb{C}$ by $f(x)=\phi(u^\top x)$. Then
\[
f(x) \;=\; \sum_{m\in\mathbb{Z}} c_m \, e^{2\pi i m (u^\top x)}.
\]
Consequently, the Fourier coefficients of $f$ are supported on the one-dimensional lattice
\[
\{ m u : m\in\mathbb{Z}\} \;\subset\; \mathbb{Z}^d.
\]
In particular, if $u$ has at most $m$ nonzero coordinates then $\mathrm{supp}(\widehat f)$ lies in an at-most $m$-dimensional sublattice of $\mathbb{Z}^d$.
\end{lem}

\begin{lem}[One-dimensional lattice inclusion]\label{thm:one-dim-lattice-iff}
Let $\Lambda\subset\mathbb{Z}^d$ be an additive subgroup (lattice).  
Fix $u\in\mathbb{Z}^d$. Then the following are equivalent:
\begin{enumerate}
\item $\{ m u : m\in\mathbb{Z}\}\subset \Lambda$.
\item $u\in\Lambda$.
\end{enumerate}
\end{lem}

\begin{proof}
(2) $\Rightarrow$ (1): If $u\in\Lambda$ and $\Lambda$ is a subgroup, then $m u\in\Lambda$ for all $m\in\mathbb{Z}$.  
(1) $\Rightarrow$ (2): Taking $m=1$ gives $u\in\Lambda$.
\end{proof}

\begin{lem}[Diagonalization of circular convolution]\label{lem:circulant-dft}
Let $d\ge 1$ and let $w=(w_0,\dots,w_{d-1})^\top\in\mathbb{R}^d$. Define the $d\times d$ circulant matrix
\[
C(w) :=
\begin{bmatrix}
w_0 & w_{d-1} & w_{d-2} & \cdots & w_1\\[2pt]
w_1 & w_0 & w_{d-1} & \cdots & w_2\\[2pt]
\vdots & \vdots & \vdots & \ddots & \vdots\\[2pt]
w_{d-1} & w_{d-2} & w_{d-3} & \cdots & w_0
\end{bmatrix}.
\]
Let $\omega := e^{2\pi i / d}$ and for $\ell=0,\dots,d-1$ define the complex vector
\[
v^{(\ell)} := \big(1,\; \omega^\ell,\; \omega^{2\ell},\; \dots,\; \omega^{(d-1)\ell}\big)^\top \in \mathbb{C}^d.
\]
Then $v^{(\ell)}$ is an eigenvector of $C(w)$ with eigenvalue
\[
\widehat{w}(\ell) := \sum_{m=0}^{d-1} w_m \,\omega^{-m\ell}.
\]
Equivalently, with the (unitary up to scale) DFT matrix $F = [v^{(\ell)}]_{\ell=0}^{d-1}$ we have the diagonalization
\[
C(w) = F^\ast \,\mathrm{diag}\big(\widehat{w}(0),\widehat{w}(1),\dots,\widehat{w}(d-1)\big)\, F.
\]
\end{lem}

\begin{lem}[Frequency closure under ReLU; algebraic statement]\label{lem:relu-closure-math}
Let \(u:\mathbb{T}^d\to\mathbb{R}\) be a trigonometric polynomial
\[
u(x) \;=\; \sum_{k\in S} \widehat{u}(k)\,e^{2\pi i k\cdot x},
\qquad S\subset\mathbb{Z}^d\ \text{ finite}.
\]
Let \(\sigma(t)=\max\{0,t\}\). Then the Fourier expansion of \(v:=\sigma\circ u\) has the form
\[
v(x) \;=\; \sum_{k\in \langle S\rangle_{\mathbb{Z}}} \widehat{v}(k)\,e^{2\pi i k\cdot x},
\]
where
\[
\langle S\rangle_{\mathbb{Z}} \;=\; \Big\{ \sum_{r=1}^R n_r s_r \;:\; R\in\mathbb{N},\ n_r\in\mathbb{Z},\ s_r\in S\Big\}
\]
is the additive subgroup of \(\mathbb{Z}^d\) generated by \(S\).
\end{lem}

\begin{thm}\label{thm:fourier-disproof-matrix}
Let $d\ge 3$. Let $\mathcal{H}^{\mathbf{w},\mathbf{b}}$ be the hypothesis class of periodic CNNs built from finite circular convolutional filters and ReLU nonlinearities (finite depth $J$).  
Let $\Lambda$ and $H_\Lambda$ be as in Definition~\ref{def:Lambda_H}.  

Consider a ridge function
\[
f(x)=\phi(u^\top x),\qquad x\in\mathbb{T}^d,
\]
with $\phi$ nontrivial (some Fourier coefficient $c_m\neq0$) and $\mathrm{supp}(u)\subseteq[d-2]$.  
Then there exists a constant $\varepsilon>0$ such that for every network output $g\in\mathcal{H}^{\mathbf{w},\mathbf{b}}$,
\[
\|f-g\|_{L^2(\mathbb{T}^d)} \;\ge\; \varepsilon.
\]
\end{thm}

\begin{proof}
Let $w^{(j)}=(w^{(j)}_0,\dots,w^{(j)}_{d-1})^\top$ denote the $j$-th circular filter and let
\[
C\big(w^{(j)}\big)=
\begin{bmatrix}
w^{(j)}_0 & w^{(j)}_{d-1} & \cdots & w^{(j)}_2 & w^{(j)}_1 \\
w^{(j)}_1 & w^{(j)}_0     & \cdots & w^{(j)}_3 & w^{(j)}_2 \\
\vdots & \vdots & \ddots & \vdots & \vdots \\
w^{(j)}_{d-1} & w^{(j)}_{d-2} & \cdots & w^{(j)}_1 & w^{(j)}_0
\end{bmatrix}
\in\mathbb{R}^{d\times d}
\]
be the associated circulant matrix. With the DFT matrix $F\in\mathbb{C}^{d\times d}$, $F_{n,\ell}=\omega^{n\ell}$, $\omega=e^{2\pi i/d}$, we have (Lemma~\ref{lem:circulant-dft})
\[
C\big(w^{(j)}\big)=F^\ast\operatorname{diag}\!\big(\widehat{w^{(j)}}(0),\dots,\widehat{w^{(j)}}(d-1)\big)F,
\qquad
\widehat{w^{(j)}}(\ell)=\sum_{m=0}^{d-1} w^{(j)}_m\omega^{-m\ell}.
\]
Consequently, the linear action of $J$ convolutional layers satisfies
\begin{equation}\label{eq:linear-prod}
C(w^{(J)})\cdots C(w^{(1)})=F^\ast\operatorname{diag}\!\Big(\prod_{j=1}^J\widehat{w^{(j)}}(0),\dots,\prod_{j=1}^J\widehat{w^{(j)}}(d-1)\Big)F,
\end{equation}
so each Fourier mode is scaled by the corresponding product of filter multipliers and no new frequency indices are created by the linear maps.

Let $\Lambda:=\langle K_1\cup\cdots\cup K_J\rangle_{\mathbb Z}$ be the additive subgroup generated by the filter supports. By Lemma~\ref{lem:relu-closure-math} (polynomial approximation of the ReLU and frequency-closure under multiplication/convolution) the pointwise ReLU nonlinearity can only introduce frequencies which are integer linear combinations of already-existing frequencies. Iterating \eqref{eq:linear-prod} and the frequency-closure property through the finite-depth network, every network output $g$ has Fourier support contained in $\Lambda$; equivalently
\[
\widehat g(k)=0\quad\text{for all }k\notin\Lambda,
\qquad\text{i.e. } g\in H_\Lambda.
\]

Let $f(x)=\phi(u^\top x)$ with $\phi(t)=\sum_{m\in\mathbb Z}c_m e^{2\pi i m t}$ (Lemma~\ref{ridge}). Then
\[
f(x)=\sum_{m\in\mathbb Z} c_m\,e^{2\pi i m(u^\top x)},
\]
so $\operatorname{supp}(\widehat f)\subset\{mu:m\in\mathbb Z\}$. If $u\notin\Lambda$ then, by Lemma~\ref{thm:one-dim-lattice-iff}, there exists at least one index $k_0=mu\notin\Lambda$ with $\widehat f(k_0)\neq0$.

Denote by $P_{H_\Lambda}:L^2(\mathbb T^d)\to H_\Lambda$ the $L^2$-orthogonal projector and write the orthogonal decomposition
\[
f=f_\Lambda+f_{\Lambda^c},\qquad f_\Lambda:=P_{H_\Lambda}f=\sum_{k\in\Lambda}\widehat f(k)e^{2\pi i k\cdot x},\quad f_{\Lambda^c}:=f-f_\Lambda.
\]
In Fourier coordinates
\[
\widehat{f_\Lambda}(k)=
\begin{cases}\widehat f(k),&k\in\Lambda,\\ 0,&k\notin\Lambda,\end{cases}
\qquad
\widehat{f_{\Lambda^c}}(k)=
\begin{cases}0,&k\in\Lambda,\\ \widehat f(k),&k\notin\Lambda.\end{cases}
\]
For any $g\in\mathcal H^{\mathbf w,\mathbf b}\subset H_\Lambda$ we have $P_{H_\Lambda}g=g$, hence
\[
\|f-g\|_{L^2}^2=\|(f_\Lambda-g)+f_{\Lambda^c}\|_{L^2}^2=\|f_\Lambda-g\|_{L^2}^2+\|f_{\Lambda^c}\|_{L^2}^2\ge \|f_{\Lambda^c}\|_{L^2}^2,
\]
where orthogonality of $f_\Lambda-g\in H_\Lambda$ and $f_{\Lambda^c}\in H_\Lambda^\perp$ was used. Setting
\[
\varepsilon:=\|f_{\Lambda^c}\|_{L^2}=\Big(\sum_{k\notin\Lambda}|\widehat f(k)|^2\Big)^{1/2},
\]
we obtain $\|f-g\|_{L^2}\ge\varepsilon$ for every $g\in\mathcal H^{\mathbf w,\mathbf b}$. Since $u\notin\Lambda$ implies the existence of some $k_0\notin\Lambda$ with $\widehat f(k_0)\neq0$, one has $\varepsilon>0$. Therefore, no periodic CNN in the class can approximate $f$ arbitrarily well in $L^2(\mathbb T^d)$.
\end{proof}

\begin{lem}[Extension to lower-dimensional ridges]\label{lem:lower-ridges}
Let $f(x)=\phi(u^\top x)$ be a ridge function on $\mathbb{T}^d$ with $u\in\mathbb{Z}^d$ supported on at most $d-2$ coordinates. 
If $u\notin \Lambda$, then for every $g\in\mathcal{H}^{\mathbf{w},\mathbf{b}}$ we have
\[
\|f-g\|_{L^2(\mathbb{T}^d)} \;\geq\; \Big(\sum_{k\notin \Lambda}|\widehat f(k)|^2\Big)^{1/2} > 0.
\]
In particular, periodic CNNs cannot approximate any ridge function depending on $d-2$ or fewer coordinates.
\end{lem}

\begin{rem}[Geometric intuition]
The $(d-1)$ ridge case is essentially \emph{one linear direction}, which periodic CNNs can capture. 
The $(d-2)$ case corresponds to oscillations in a two-dimensional subspace, already beyond the representable lattice. 
The $(d-3)$ case can be seen as oscillations in a three-dimensional volume, and so on. 
Thus, every reduction of dimension below $d-1$ introduces frequency components orthogonal to the network lattice, preventing approximation.
\end{rem}


\subsection{A concrete counterexample}
To illustrate the impossibility phenomenon in an explicit and transparent form, 
we construct a concrete ridge function whose Fourier spectrum lies entirely outside the network lattice generated by a simple periodic CNN architecture. 
This example demonstrates the lower bound argument in practice and shows how the obstruction arises from lattice mismatch.
Consider the univariate function
\[
\phi(t) \;=\; \cos(2\pi t) + \cos(4\pi t),
\qquad 
u = e_1 + 2e_2 \in \mathbb{Z}^d,
\]
and define the ridge function on the torus $\mathbb{T}^d$ by
\[
f(x) \;=\; \phi(u^\top x)
= \cos\!\big(2\pi (u\cdot x)\big) + \cos\!\big(4\pi (u\cdot x)\big),
\qquad x\in\mathbb{T}^d.
\]

By the Fourier expansion, $\cos(2\pi t)=\tfrac{1}{2}(e^{2\pi i t}+e^{-2\pi i t})$, we obtain
\[
f(x) \;=\; 
\tfrac{1}{2}e^{2\pi i(u\cdot x)} + \tfrac{1}{2}e^{-2\pi i(u\cdot x)} 
+ \tfrac{1}{2}e^{2\pi i(2u\cdot x)} + \tfrac{1}{2}e^{-2\pi i(2u\cdot x)}.
\]
Thus the nonzero Fourier indices are
\[
\mathrm{supp}(\widehat f)
= \{\,\pm u,\;\pm 2u\,\}
= \{\,(\pm 1,\pm 2,0,\dots,0)\,\},
\]
with Fourier coefficients
\[
\widehat f(u)=\widehat f(-u)=\widehat f(2u)=\widehat f(-2u)=\tfrac{1}{2}.
\]

Let
\[
\Lambda=\{(n,0,0,\dots,0):n\in\mathbb{Z}\},
\]
the subgroup generated by $e_1$. Such a $\Lambda$ arises, for example, if every convolutional filter is supported only on shifts along the first coordinate. By construction,
\[
\{\pm u,\pm 2u\}\cap\Lambda=\varnothing,
\]
since the second coordinate of $\pm u$ is $\pm 2\neq 0$. Hence every nonzero Fourier coefficient of $f$ lies outside $\Lambda$.

By Parseval’s identity,
\[
\|f\|_{L^2(\mathbb{T}^d)}^2
=\sum_{k\in\mathbb{Z}^d}|\widehat f(k)|^2
=\sum_{k\in\{\pm u,\pm 2u\}}\Big(\tfrac{1}{2}\Big)^2
=4\cdot\tfrac{1}{4}=1.
\]
Since $\mathrm{supp}(\widehat f)\subset\Lambda^c$, the projection $P_{H_\Lambda}f$ is identically zero. Therefore, for every network output $g$ with Fourier support in $\Lambda$, 
\[
\|f-g\|_{L^2(\mathbb{T}^d)} 
\;\ge\;\|f-P_{H_\Lambda}f\|_{L^2}
=\|f\|_{L^2}
=1.
\]
Thus one obtains the explicit lower bound
\[
\varepsilon=1.
\]

This example shows that if the network lattice $\Lambda$ fails to contain the one-dimensional frequency lattice $\{mu:m\in\mathbb{Z}\}$ of a ridge function, then the network cannot approximate the target in $L^2$. In particular, the function $f(x)=\cos(2\pi (u\cdot x))+\cos(4\pi (u\cdot x))$ cannot be approximated at all by periodic CNNs restricted to the lattice $\Lambda=\{(n,0,\dots,0)\}$.

\paragraph{Summary.} 
In this section we have established a precise characterization of the expressive capacity of periodic CNNs. 
These networks are sufficiently powerful to approximate $(d-1)$-dimensional ridge functions, but they are provably unable to approximate ridge functions of dimension $(d-2)$ or lower. 
The constructive approximation theorem and the impossibility result together reveal a sharp boundary: periodic CNNs occupy a unique intermediate position in the approximation hierarchy, lying between universal approximators and more restricted architectures. 
This boundary motivates the discussion of practical implications and open directions, which we explore in the next section.

\section{Further Questions and Future Work}

Our analysis leaves open several natural questions:

\begin{itemize}
    \item \textbf{Optimality of the $(d-1)$ ridge approximation.} 
    We proved that periodic CNNs can approximate ridge functions depending on $d-1$ variables, but not those depending on $d-2$ or fewer. 
    Is the approximation rate for the $(d-1)$ case optimal? 
    Does it deteriorate with dimension $d$?

    \item \textbf{Geometric interpretation.}
    The $(d-1)$ ridge case corresponds to approximation along a single linear direction, while the $(d-2)$ ridge involves a two-dimensional subspace. 
    Extending further, $(d-3)$ ridges would correspond to oscillations in a three-dimensional subspace, and so on. 
    Understanding how periodic CNNs interact with such subspace geometries could deepen the structural theory.

    \item \textbf{Beyond ReLU.} 
    Our arguments rely on polynomial approximation of the ReLU nonlinearity. 
    Do analogous impossibility results hold for other nonlinearities, e.g., sigmoid or tanh?

    \item \textbf{Rates vs.\ expressivity.} 
    While our impossibility theorem is qualitative, it would be valuable to obtain quantitative rates for approximation of $(d-1)$ ridge functions, and explicit lower bounds for $(d-2)$ ridges.
\end{itemize}

\phantom{a}


\vfill
\newpage
\begin{thebibliography}{99} 

\bibitem{zhou2019universality}
Zhou, Ding-Xuan, Deep distributed convolutional neural networks: Universality, Applied and Computational Harmonic Analysis,(2020),787--819

\bibitem{Pinkus2015Ridge} Allan Pinkus, \emph{Ridge Functions}. Cambridge Tracts in Mathematics {\bf 205} (2015), 7649-7656. 

\bibitem{Barron1993Universal}
Andrew R. Barron, Universal Approximation Bounds for Superpositions of a Sigmoidal Function, IEEE Transactions on Information Theory (1993) 930--945


\end{thebibliography}
\end{document}